\documentclass[letterpaper, 10 pt, conference]{IEEEtran}

\IEEEoverridecommandlockouts

\usepackage[english]{babel}
\usepackage[caption=false,font=footnotesize]{subfig}
\usepackage{cite}
\usepackage{amsmath,amssymb,amsfonts}
\usepackage{graphicx}
\usepackage{textcomp}
\usepackage{xcolor}
\usepackage{caption}

\usepackage{rotating}
\usepackage{tabularx}
\usepackage[labelfont=bf]{caption} 
\usepackage{ragged2e}
\usepackage{lipsum}

\usepackage{booktabs, makecell, tabularx}
\usepackage{stfloats}
\usepackage{siunitx}
\usepackage{multicol}
\usepackage{authblk}

\usepackage{hyperref}
\usepackage{graphicx}

\hypersetup{
    colorlinks=true,
    linkcolor=blue,
    filecolor=magenta,      
    urlcolor=blue,
    pdftitle={Overleaf Example},
    pdfpagemode=FullScreen,
    }

\def\BibTeX{{\rm B\kern-.05em{\sc i\kern-.025em b}\kern-.08em
    T\kern-.1667em\lower.7ex\hbox{E}\kern-.125emX}}
\begin{document}

\title{Roadmap to Autonomous Surgery - A Framework to Surgical Autonomy}


\author[1]{Amritpal Singh,  \thanks{Corresponding author: \href{mailto:}{amrit@gatech.edu}  \href{https://orcid.org/0000-0002-3229-9460}{\includegraphics[width=0.17in,height=0.17in]{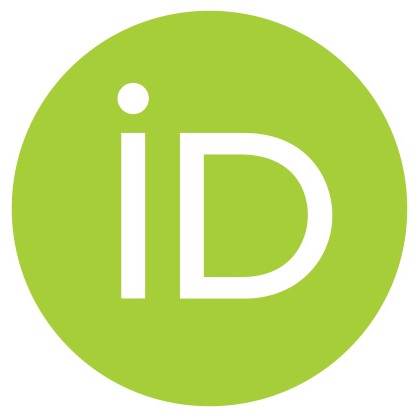}}  }}
\affil[1]{
College of Computing \\ 
Georgia Institute of Technology, USA
}

\maketitle

\begin{abstract}

Robotic surgery has increased the domain of surgeries possible. Several examples of partial surgical automation have been seen in the past decade. We break down the path of automation tasks into features required and provide a checklist that can help reach higher levels of surgical automation. Finally, we discuss the current challenges and advances required to make this happen. 

\end{abstract}


\section{Introduction and Motivation}
The term "Autonomous surgery" has been used repeatedly in the literature [1] [2] [3] [4] [5] [6] [7] with different interpretations in the author's mind. It has been used for pure robotic surgery without autonomous control, training resident doctors, suturing automation, and path planning.

Robotic surgery systems like Da Vinci surgery robots provide a mechanical system to develop and test different automation efforts. In general, Robotic surgeries offer several advantages over minimal-invasive surgeries -  perception augmentation( medically imaging - allowing better surgical plan execution, avoiding accidental trauma to vital organs, ability to not miss behind any tumor region. and manipulation augmentation(overcoming technical difficulties of using a manual laparoscopic tool - allowing increased distal dexterity, manipulation of multiple instruments, improved precision, and steadiness, allowing multiple surgeons to manipulate multiple instruments and collaborate on surgeries). This will also allow surgeons to have a lower cognitive and physiological burden associated with minimal instrumental laparoscopy.

While previous works [1] [2] [8] [9] have laid the foundations on different levels, the description of tasks required to achieve them is not discussed clearly. Since several technological breakthroughs will be required in robotics, artificial intelligence, and other industries, rather than talking about these relevant technologies, we will talk about checkpoints required to be achieved. In this article, we talk about sub-components of these systems that will achieve a higher level of automation in different surgical components. We will talk about why thinking in this framework is essential and its potential benefits. This work aims to lay a roadmap for robotics and AI engineers to get to higher levels of surgical automation. 

\begin{figure}
    \centering
    \includegraphics[width=0.45\textwidth]{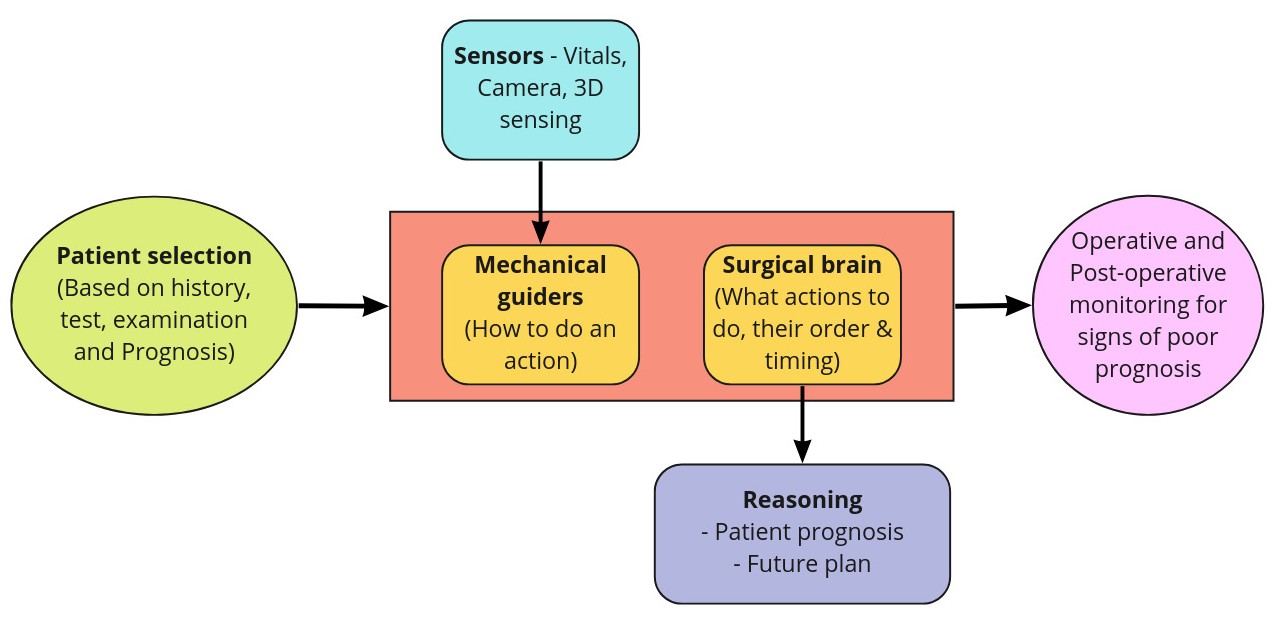}
    \caption{Different points of interactions}
    \label{interactions}
\end{figure}

\section{Breakdown of a typical surgery}

A typical surgery involves many dynamics and is a team effort. The team generally consists of at least a surgeon, nurse, and anesthetist. There is an entire sequence of tasks done in any regular surgery. While the details might vary based on site/type of surgery, the main flow stays somewhat similar. Table.~\ref{tablecomponents}  represents different phases of surgery. Of these, automated systems can be relatively good at postoperative tasks like documentation. At the same time, anesthesia inducing and waning off is a difficult task in itself. For full phase 5 in scenarios like military, battle surgery, or deep space exploration, this becomes more critical, as there might not be someone to assist with these tasks.

\begin{table*}[h!]
\centering
\begin{tabular}{ |c|c|c|c| }
 \hline
 Phase     &      Surgeon      & Nurse  &   Anaesthetist  \\
 \hline\hline
 Pre-operative planning      &   Imaging/Tests required, Equipment planning  & Sterilise instruments & Induce anaesthesia   \\
 Intra-operative prep      &  Locate entry point, surgical scrub over patient  & Assist & Monitoring vitals  \\
 Puncture      &      Skin incision, Port creation  & Assist & Monitoring vitals  \\
 Operative phase  &    Perform surgery  & Assist & Monitoring vitals \\
 Concluding steps  &    Skin suturing  & Assist & reduce anaesthesia \\
 Post-operative      &    Documentation  & Documentation  & Patient hand over  \\
 \hline
\end{tabular}
\caption{Dynamics of typical surgery}
\label{tablecomponents}
\end{table*}

\subsection{Phases of surgery}

Any surgery can be divided mainly into preoperative, operative, and postoperative phases. 
The perioperative period starts when the patient is informed of the need for surgery until the patient resumes their normal activities after the surgical procedure. The preoperative phase starts after the patient is informed of the need for surgery and is prepared physically and psychologically for the surgery. Intra-operative phase mainly spans the patient's stay inside the operative room. Postoperative phases begin from the patient's transfer from the operative room until the resolution of surgery sequelae.
 
In 2017, authors in [2] introduced the six levels of surgical automation, starting from no automation at level 0 to full automation at level 5. [1], [2] elaborated on this definition with examples of some commercially available versions of them. Currently, most approved systems are level 1, with few examples of level 2 and 3 systems. Rising higher levels will help augment surgeons, allowing higher precision surgery and improvements in time efficiency. Levels 4 and 5 systems will be capable of automating some or most parts of surgery. These systems are primarily fictional as of today and can have a potential use case to treat soldiers in battle and deep space explorations. Substantial efforts are being put into bringing this into reality. These frameworks do not consider preoperative and postoperative care, which play an essential role in patient prognosis and are critical to developing level 5 systems. 

In this work, we introduce different wings of surgical automation - Preoperative, Operative, and Postoperative. Most work has happened in the pre and postoperative phases. Over the last decade, we have started seeing more and more work happening around the operative phase - suturing, path planning, and a better field of view. 

\begin{figure}
    \centering
    \includegraphics[width=0.3\textwidth]{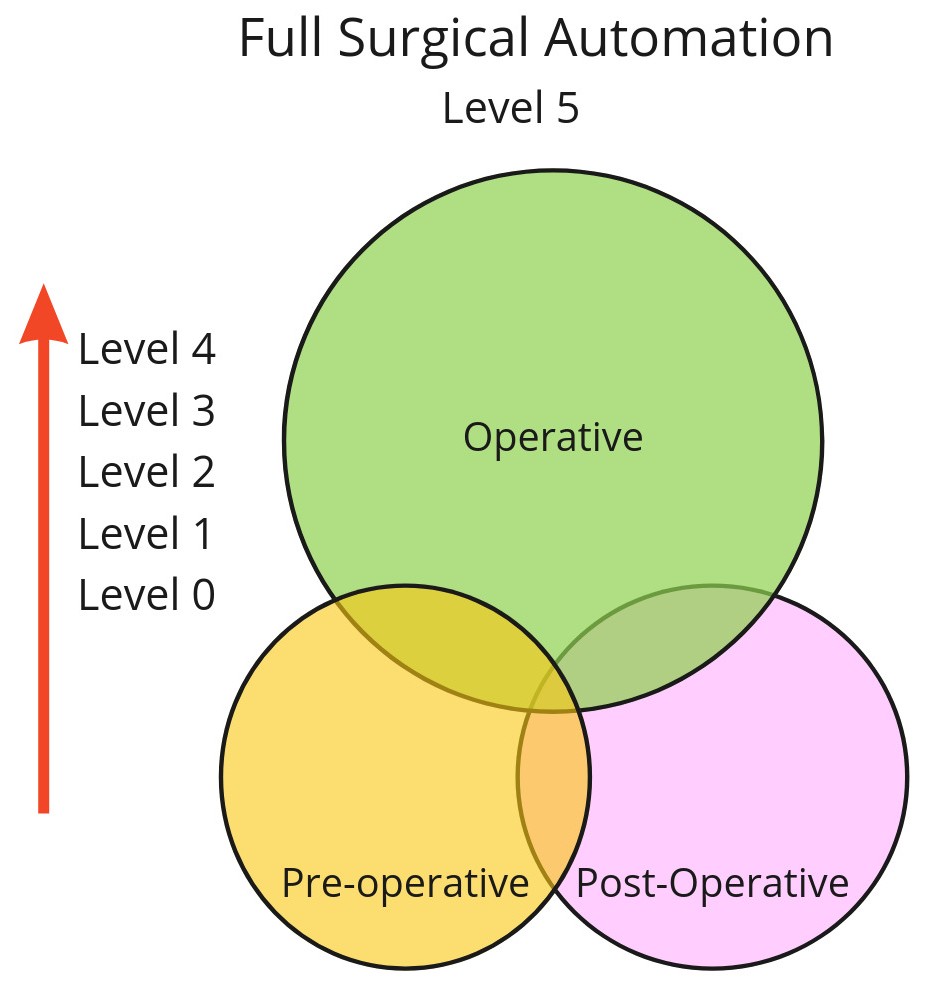}
    \caption{Different levels of Surgery automation}
    \label{levels}
\end{figure}

\subsubsection{Preoperative surgery}
A lot of work has happened around the preoperative phase. This ranges from patient selection for surgery, prognosis prediction, patient prep for surgery, instrument planning, and scrubbing patients. The last 2 need more attention. 

\subsubsection{Operative surgery and Levels of Automation}

Other than performing physical surgical procedures, any autonomy will require patient monitoring along with other well-established requirements like navigating to the area of interest, tissue manipulation, and Dissect/Implant/Incision. Following is a summary of levels of automation as discussed in several works [3] [4] [5] [6] [7].

\begin{itemize}
\item  Level 0 - No automation - Manually controlled. Humans monitor patient vitals. Open surgeries and Mechanically lap surgeons are examples of this. The human surgeon performs all the tasks
\item Level 1 - Surgical assistance - Humans are assisted but still physically perform the surgery. Intraoperative image guidance.
\item Level 2 - Partial automation - Robotic hip arthroplasty robot. - reduced level of human input required. Predict emergencies, Patient prognosis, Efficiency - Helps in - Navigation, difficult vision, saves time in long procedures
Predict emergencies, Patient prognosis 
\item  Level 3 - Conditional Automation - Automated some components. Cyberknife, Automated preoperative planning. Surgeons provide instruction to execute a specific task.

\item  Level 4 - High Automation
Robots are capable of performing most, if not all, parts of surgery. While the system could do many of the sub-tasks like - Initial puncture, Instrument navigation, Exploration of field, manipulation of field, dissection/resect/implant the region, These actions are performed only under human supervision, with the ability to override the actions. Since these systems will require monitoring patients - the ability to monitor and respond to patient vitals will be required. Will need to require detecting/predicting emergencies, with follow-up response to restrict damage or request override.
\item  Level 5 - Full Automation - Do not require human assistance/ monitoring - Dynamic surgical task. Fully autonomous, versatile. It can be used in space exploration. These systems could potentially innovate new, more efficient ways of surgery. Human override components can be removed here. These systems can do all the sub-tasks as well - Initial puncture, Instrument navigation, Exploration of field, manipulation of field, dissection/resect/implant of the region, as well as plan for these actions.
\end{itemize}

Level 5 systems will be a great step of artificial narrow intelligence, with the ability to do specific complex tasks, with complex planning in closed environments. Achieving this will take us one step closer to our understanding to build AGI.

\subsubsection{Postoperative surgery}

Attempts have been made to predict postoperative complications [37] and adverse events [38] using NLP methods. Postoperative phase patients need to be dosed off from the anesthesia and require active monitoring. This will be crucial to no table-side surgeon/physician.

\begin{table*}[ht]
    \centering
    \begin{tabular}{ | p{0.07\linewidth} | p{0.1\linewidth} | p{0.35\linewidth} | p{0.35\linewidth} |}
    
    \hline
     Levels     &      Task      & Human interaction/ monitoring &   Features  \\
     \hline\hline
     Level 0      &    No Autonomy  & exclusively surgeon  &   None  \\
     Level 1      &    Robot Assistance  & exclusively surgeon + supportive robot   &   Tool tracking, Eye tracking, Stabilisation, Haptic feedback  \\
     Level 2      &    Task Autonomy  & perform task, but only after all parameter of task are defined  &   Suturing, Tissue retraction, Ablation  \\
     Level 3      &    Conditional Autonomy  & devices strategy, to be approved by humans  &   Tissue modelling, Navigation\\
     Level 4      &    High Autonomy  & make a decision and execute it autonomously, under supervision of surgeon &  Tissue segmentation, resection, debridement\\
     Level 5      &    Full Autonomy  & Perform all procedure on its own  &   -   \\
    \hline
    
    \end{tabular}
    \caption{Different levels of Automation for Operative phases}
    \label{tab:automation}
\end{table*}

\section{Current level of automation}

With the development of robotic surgery, even without any automation (level 0), we can perform tele-operative surgeries. This became more important during COVID times when elective surgeries could be performed without risk of contamination. Adding a layer of augmentation or automation to these robotic systems will help increase surgical efficiency. To name a few - Increased efficiency and precision, hence reducing the duration of surgeries and making longer-duration surgeries feasible. Augmenting human surgeons will allow for Intelligent maneuvers, tissue damage avoidance, suturing in hard-to-reach areas, camera control [10], Disturbance rejection to minimize tissue damage [11], and path planning in soft tissue surgeries [12]. Authors in [13] proposed another system for Smart Tissue Autonomous Robot (STAR) to enable semi-autonomous robotic anastomosis on deformable tissue. 

With the expansion of universal healthcare access to surgery, there is going to be a deficiency of surgeons to tackle these tasks. High-Income countries have 36\% surgeons while their population is 18\% of the world. The low-income countries that account for 12\% world population get only 2\% of surgeons[35][36]. The estimated increase in surgeons will not catch up with the required demand to meet unmet surgical needs. Increasing the augmentation of surgeons can help tackle this to some extent.

\subsection{Assistive tasks}
Autonomous robotic systems for the removal of blood during surgical procedures have been developed. In one such example, a 6 DOF robot with a dual camera, aspirator, and a path planning algorithm to reach the aspiration site. [14].

In other cases, an algorithm for autonomous grasping of grasp soft tissues in different modes is damage, slip, or safe grasp. On testing, no signs of tissue tear or slippage were found[15]. The HEARO robotic system is another surgical robot to assist the surgeon with orientation, reference information to anatomical structures, and drill trajectory during otological and neurosurgical procedures. [16]

\subsection{Partial Automation}
Collaborative systems can help increase the efficiency of several surgical processes. One such system was developed using a Smart tissue Autonomous robot(STAR) for confidence-based supervised autonomous suturing to perform robotic suturing tasks along with a surgeon collaboratively. Suture placement accuracy was 94.74\% on pure STAR automation and 98.1\% accuracy with a 25\% human intervention[17].

Autonomous suturing has also been tried for in-vivo open soft tissue surgery, using plenoptic three-dimensional and near-infrared fluorescent (NIRF) imaging system. Consistency of suturing was tracked based on their average suture spacing, the number of mistakes that required removing the needle from the tissue, completion time, pressure at which the anastomosis leaked, and lumen reduction in intestinal anastomosis. In comparison to manual laparoscopic surgery and clinically (using RAS approaches), the outcome of autonomous procedures was found superior to surgery performed by expert surgeons. The system received this despite dynamic scene changes and tissue movement during surgery[18]. 

 present the Dexterous Surgical Skill (DESK) database for 
Knowledge transfer between robots has also been tried. Two such approaches - No transfer and domain transfer have been compared. In the no-transfer scenario, the training and testing data were obtained from the same domain, compared to a domain-transfer scenario where training data is a blend of simulated and real robot data but tested on real robot data only. Authors in [19] tested this on several types of robots and found that the transfer model showed an accuracy of 81\%, 97.5\%, and 93\% for the YuMi, Taurus II, and the da Vinci robot, respectively. While, in the YuMi case, the ratio of real-to-simulated data was 22\% to 78\%, later were trained only with simulation data. This shows great promise for the augmentation and automation of sub-tasks.

\begin{figure}
    \centering
    \includegraphics[width=0.4\textwidth]{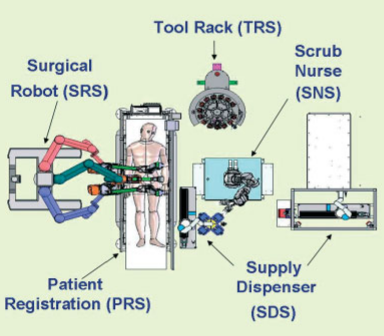}
    \caption{Trauma pod - diagram replicated from [20] }
    \label{trauma_pod}
\end{figure}

\subsection{Limitations}

The limitation of such systems will be the ability to help in robotic surgeries only. Open surgeries cannot be augmented. 
On the other hand, there are surgeries that are not performed by robotic surgery systems like Da Vinci. One such standard procedure is a cesarean section. At times, some laparoscopic surgeries are also converted into open surgeries. Doing so with a fully autonomous system could be challenging.

\section{Components of Surgical Automation}

\begin{itemize}
\item Instruction engine - responsible for gathering instructions about the current procedure to perform/ assist. The interface that can support this can be the screen, gesture/button, and voice-based. 
Devices already exist and are used today that are voice-activated (ViKY, Endocontrol, Grenoble, France), and it is conceivable that NLP could eventually evolve to benefit the action of voice-controlled devices during a procedure [21], [22]

\item Execution engine - Executing actions with precision after being told. Variety of tasks to solve for - Suturing, Exploration, Manipulation, Incision, Drainage, Cauterise, Pick and Place, Implant. This area has received recently improved due to deep learning-based methods.

\item Planning and coordination engine - Create a standard plan to control a local procedure to some or full extent, given ideal situations. Decides what to do, when to do, and when to transition between sub-tasks. Works on a relatively long time horizon.

\item Navigation engine - Extension to exploration tasks. Involves local navigation to reach a site of interest. (e.g. mechanical thrombectomy. Adaptive human anatomy understanding

\item Emergency engine - This allows the system to predict, detect, react and control any emergency conditions. Doing so requires differentiating between planned vs unplanned events, tolerable, unplanned, and unwanted events. Overall, this requires looking at a longer horizon to predict such events but also being able to act on them in a smaller time horizon.

\item Self-doubt engine- This acts as an internal safety system to raise flags about what it can’t handle (“I don’t know this” system) and raises appropriate warning/human override requests. 

\item Observer engine - This will allow observer ability like humans, hence imitate behaviors given adequate information and patient parameters. Allows instruction fewer actions.

\item Innovator engine - This allows the system to innovate more efficient approaches to perform previous tasks efficiently or level up to perform tasks previously not feasible. This requires the system to have the ability to measure what to improve and perform controlled testing/simulation.
\end{itemize}

\begin{figure*}
        \centering
        \includegraphics[width=0.99\textwidth]{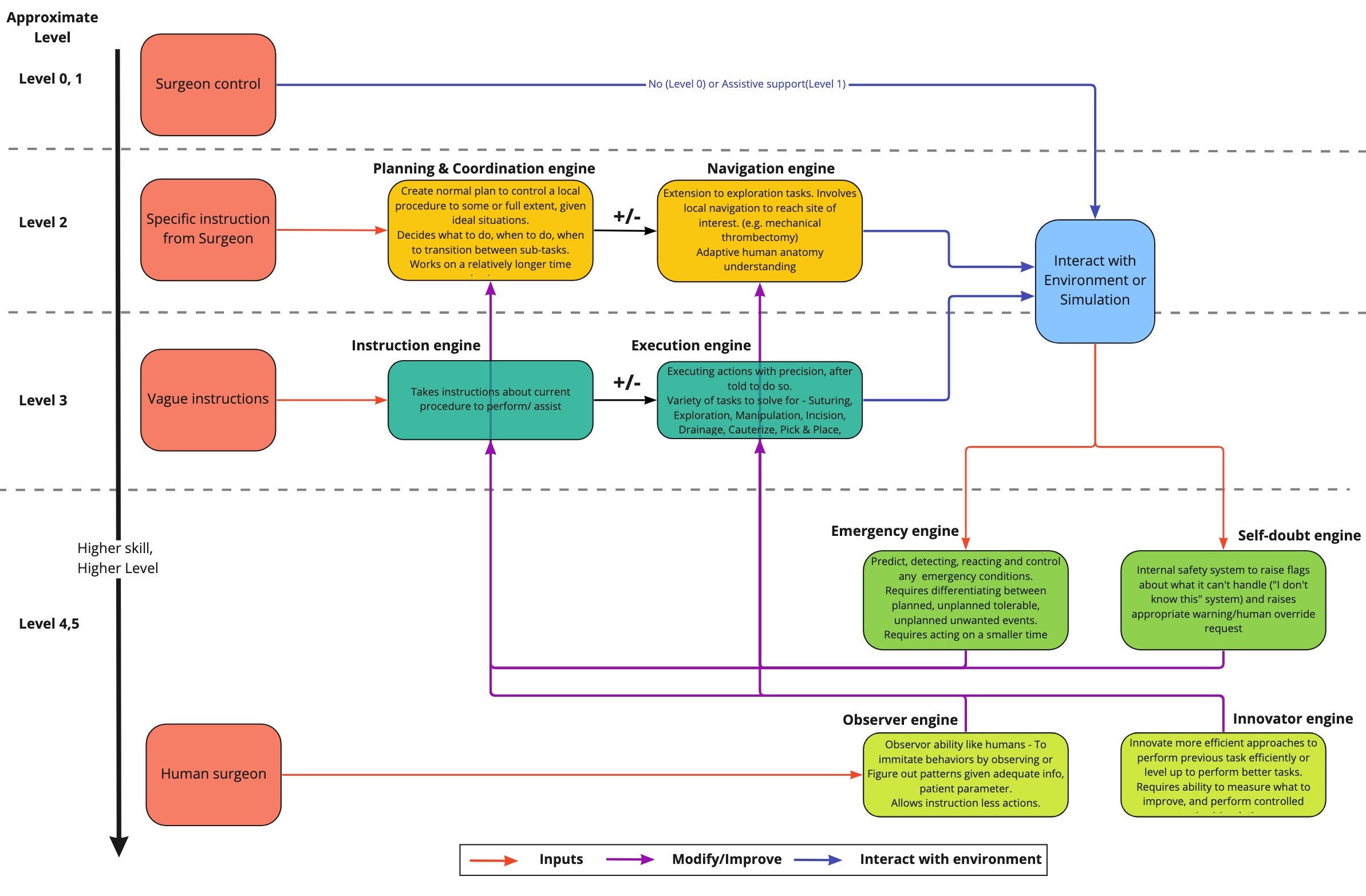}
        \caption{Different components of automated surgical systems}
        \label{components}
\end{figure*}

It is important to note that while Fig.~\ref{components} separate components as separate compartments, it is purely for representational ease. As we have seen in previous technology cycles(audio detection systems), models get more straightforward with minimal sub-components as our understanding of a task improves. Our hope is to see more such examples in surgery automation. 

\section{Challenges and area of improvements}

Robotic surgical systems can help to increase OR autonomy through camera control, application of intelligent instruments, and even accomplishment of automated surgical procedures. Currently, the autonomy level of robots depends on three factors: the complexity of the task, the environment in which the robot operates, and the required level of human-robot interaction[23]. 

Automation of robotic surgery will inherently carry all the challenges posed to robotic surgery - Namely - Higher cost, disposable instruments, and annual service charges. Others include lack of haptics or tactile feedback, fixed positioning of the operating table, and longer operative time compared to open surgery. Transmission delay over long distances can be a significant challenge for remote control.

Another important aspect is the adaption of the general technical OR environment. This includes adaptive OR setting and context-adaptive interfaces, automated tool arrangement, and optimal visualization.[5] Integration of peri- and intraoperative data consisting of the electronic patient record, OR documentation and logistics, medical imaging, and patient surveillance data could increase autonomy. 

Authors in [24] discuss the human-machine interaction perspective about generic corporative robotic device profiles, features, and use cases. New communication channels between surgeons and robotics have been tried, for example, eye tracking and voice inputs. 
There are known frameworks for evaluating training AI[25]. Since testing robotic surgery systems in the real world is not feasible, testing frameworks are required to evaluate the surgical techniques for robotic surgeons. Surgical simulations - allows for faster experiments and testing ideas. Reinforcement learning has been used to simulate and train agents to learn the task of surgical suturing [26]  [27]  [28]. Authors in [29] used a percutaneous injection of a tumour mimic mixture for experiments on renal tumour targets. More such innovations are required. 

More detailed checkpoints are required to measure the accuracy metrics for different surgery tasks[30]. What cannot be measured cannot be improved on. A great deal of work in NLP has been possible because of a clear definition of tests/benchmarks/datasets to measure the progress. The AI community needs to agree on specific standards to evaluate such surgical systems that will allow regulatory approval. 

Finally, most work done is around Da Vinci type of robots, which have their fair share of limitations. Recently there has been a broader adaptation for Continuum robotics. Continuum robotics allow a higher degree of freedom [31]  [32] [33].
Reinforcement learning has shown great promise with self-exploration, as it eliminates the data collection process. [26]  [27]  [28]  Additional work on explainability will act as a catalyst for the initial adoption of such systems.
  
Few arguments have also been raised about whether automating the surgery should be considered a goal and propose the term "surgeon-in-the-loop".[7] Full surgical automation remains far in the future. As we pave our path towards such systems, parallel work is required to look at ethical loopholes and patient aspects. 
That being said, the enhancement of surgeons using automation is essential.

Authors in [34] showed that performing robotic major liver resection without the presence of a table-side surgeon is safe and feasible. This no-table-side surgeon will be critical for full telesurgery and any possible automation. This will also include switching instruments and patient monitoring. 
Trauma Pods [20] is a  deployable system for operative surgeries on battle/ military wounded soldiers. This can allow acute critical stabilization and/or surgical procedures, autonomously or in a tele-operative mode, on wounded soldiers on the battlefield who might otherwise die before treatment in a combat hospital could be provided. The authors also talk about the different mechanical components required to make this happen. Lessons learned here can pave the path for deeper hardware understanding for building deployable robotic surgeon pods.

\section{Conclusion}

Surgical augmentation is happening. Several systems with level 2 and 3 level automation have been developed. Although there is still a long journey ahead, in this work, we propose a framework of surgical system components required to rise higher in the level of automation. The importance of pre-operative and post-operative efforts can't be ignored, especially once we get closer to level 4 or 5.

\end{document}